\documentclass[a4paper, 10pt, conference]{ieeeconf}      % Use this line for a4 paper

\IEEEoverridecommandlockouts                              % This command is only needed if
\usepackage{graphics} % for pdf, bitmapped graphics files
\usepackage{amsmath} % assumes amsmath package installed
\usepackage{amssymb}
\usepackage{stmaryrd}
\usepackage{multirow}
\usepackage{subcaption}
\usepackage{graphicx}  % assumes amsmath package installed
\let\labelindent\relax
\usepackage{enumitem}
\usepackage{tikz}
\usepackage{pgfplots}
\usetikzlibrary{arrows,automata,intersections,positioning}
\usepackage{amsfonts}
\usepackage{dsfont}
\usepackage{cases}
\usepackage{algorithm}
\usepackage{algorithmic}
\usepackage{url}

\title{\LARGE \bf
    Kinematic Single Vehicle Trajectory Prediction Baselines\\ and Applications with the NGSIM Dataset
}

\author{Jean Mercat$^{1, 2}$, Nicole El Zoghby$^{1}$, Guillaume Sandou$^{2}$, Dominique Beauvois$^{2}$, and Guillermo Pita Gil$^{1}$\\% <-this % stops a space
\thanks{$^{1}$ Department of data fusion,
        Technocentre Renault, 78280 Guyancourt, France
        \texttt{\{Jean.Mercat,
        Nicole.El-Zoghby,
        Guillermo.Pita-Gil\}@renault.com}}
\thanks{$^{2}$ Laboratoire des signaux et des systemes,
        Centrale-Supelec, 91192 Gif sur Yvette, France
        \texttt{\{Jean.Mercat,
        Guillaume.Sandou,
        Dominique.Beauvois\}@centralesupelec.fr}}
}

\begin{document}

\tikzset{
    state/.style={
           rectangle,
           draw=black, very thick,
           minimum height=2cm,
           inner sep=2pt,
           text centered,
           }
}

\maketitle
\thispagestyle{empty}
\pagestyle{empty}

%%%%%%%%%%%%%%%%%%%%%%%%%%%%%%%%%%%%%%%%%%%%%%%%%%%%%%%%%%%%%%%%%%%%%%%%%%%%%%%%
\begin{abstract}
In the recent vehicle trajectory prediction literature, the most common baselines
are briefly introduced without the necessary information to reproduce it.
In this article we produce reproducible vehicle prediction results from simple models.
For that purpose, the process is explicit, and the code is available.
Those baseline models are a constant velocity model and a single-vehicle prediction model.
They are applied on the NGSIM US-101 and I-80 datasets using only relative positions.
Thus, the process can be reproduced with any database containing tracking of vehicle positions.
%Produced results on this database establish the three most used trajectory prediction performance indicators:
%Root Mean Squared Error (RMSE), Negative Log-Likelihood (NLL),
%and Mean Absolute Error (FDE).
The evaluation reports Root Mean Squared Error (RMSE), Final Displacement Error (FDE),
Negative Log-Likelihood (NLL), and Miss Rate (MR).
The NLL estimation needs a careful definition because several formulations that differ from the mathematical definition
are used in other works.
This article is meant to be used along with the published code to establish baselines for further work.
An extension is proposed to replace the constant velocity assumption with a learned model using
a recurrent neural network.
This brings good improvements in accuracy and uncertainty estimation and opens possibilities for both complex and
interpretable models.

\end{abstract}

%%%%%%%%%%%%%%%%%%%%%%%%%%%%%%%%%%%%%%%%%%%%%%%%%%%%%%%%%%%%%%%%%%%%%%%%%%%%%%%%
\section{INTRODUCTION}

Automation of driving tasks aims for safety and comfort improvements.
For that purpose, every Autonomous Driving (AD) system relies on an anticipation of the
traffic scene movements.
Freeway datasets NGSIM I-80 and US-101, have been extensively used for various applications, among
which trajectory prediction.
Many variations of trajectory prediction algorithms have been produced in the literature.
Since they adopt different strategies, comparing them can be difficult.
In those cases, performance evaluations of simple models are used as a reference.
The most common reference model is kinematic prediction.
The Kalman filter~\cite{Kalman1960} is used to compute the kinematic state of the vehicles
and the associated uncertainty before producing predictions.
Then its predictive step is repeatedly used to compute predictions and the associated error estimations.
On freeway situations, the assumption of constant velocity is a good fit and leads to the simplest models.
However, various choices of Kalman filter model and parameters may produce different results.
Therefor, it is important to fit them to the dataset as cautiously as the parameters of the algorithm to be compared.
In this article we present a process to fit a constant velocity Kalman filter to the NGSIM I-80 and US-101 dataset.
We obtain better results on negative log-likelihood (NLL) than published results for the same model, and
on the same data.

Similarly to the conclusions made on pedestrian motion prediction in~\cite{Scholler2019},
we show that simple baselines produce more interesting results than the ones presented in the literature.
In particular, the NLL evaluation and a standard deviation study shows that
covariance prediction from state of the art models aren't much better than the estimation of a simple Kalman filter.

As an application, we improve the covariance estimation of a machine learning model based on Long Short-Term Memory
(LSTM~\cite{Hochreiter1997}) that predicts the future trajectory of a single vehicle without interaction.
In our model, the LSTM function is used as a command prediction in the predictive step of a Kalman filter
instead of the canonical encoder-decoder architecture.
Forcing this architecture produces a physical dynamic state instead of a uninterpretable latent state.
This restricts the computation capacity without degrading the results that are actually improved.

\section{Trained Kalman filter parameters for prediction}

The sanity check model for highway trajectory prediction is a constant velocity prediction.
It seems simplistic and easy to implement but various results may be found in the literature with no explanation
of the specific implementation and parameter choice.
For instance~\cite{Deo2018} and~\cite{Xu2019} implemented two different Kalman filters that are both called
constant velocity and produce predictions with associated error covariance matrices
but no details are given to reproduce them.
Consequently, the different results that they obtain cannot be interpreted.

This section writes the Kalman filter equations and describes the process to learn its parameters.
%The noise parameters of this filter where not fitted to the database as their prediction models were.
%This means that one of the comparison model on wich they improve is actually shown as less performant as it could be
%this makes a small difference on RMSE metric but a significant one on the NLL metric.

%In, a different implementations of the constant velocity prediction is made and used with similar but
%different data, leading to different results.
%Since this is only a sanity-check baseline, no detail is given to reproduce these results.
%In~\cite{Xu2019}, another implementation of the Kalman filter is made but the NLL metric is not given.
%Their implementation uses 10Hz observations instead of the 5Hz of~\cite{Deo2018} and the evaluation sets is
%using only the US-101 part of the NGSIM dataset and not the I-80.
%However, this was tunned for best state estimation in closed loop with innovation but not for best prediction in open
%loop with no innovation.
%This results in a very low RMSE at short horizon time and comparable RMSE in longer horizon.
%No more detail is given for that variation of the constant velocity model.

\subsection{Definition of the constant velocity prediction}

Constant velocity prediction is a short expression that efficiently describes what the model does.
However, it is an abusive expression because the velocity is modeled as quasi-constant.
In fact, an acceleration is represented by a zero-mean random noise.

Several expressions of constant velocity models for Kalman filtering with positions as observations can be made.
We use the state vector $X=(x, v_x, y, v_y)^T$ because it is the simplest linear Kalman filter for a constant
velocity prediction.
However, other work might choose $X=(x, y, \theta, v)^T$,
with the equality : $(v_x, v_y) = (\cos(\theta)v, \sin(\theta)v)$.
The bicycle model, with a state vector $(x, y, \theta, v, \omega, a)$, respectively
position, heading angle, velocity, wheel angle, and acceleration describes the actual
vehicle dynamics with a no slip approximation.
The constant velocity model might have a different meaning with this last state vector because the wheel angle could
be considered constant or set to 0 along with the acceleration.
Many variations of these models may be used.
Consequently, even if it is a simplistic model, the name constant velocity and
a brief description are not enough to reproduce the results.

With our model choice, observations are sequences of vehicle positions.
At each time $t_0$, we consider a 3 seconds observation history at 5Hz, this is the past trajectory $\{(x, y)_{k}\}_{k=-14, 0}$.
The coordinate system is centered on the vehicle position at $t_0$ with a constant orientation thus $(x, y)_0 = (0, 0)$.
These observations are used sequentially to update a Kalman filter from its initial point, so it reaches a good state
estimation at $t_0$.
With our linear model, the evolution of the state $X$ from step $k$ to step $k+1$ is written as follow:
\begin{equation}
    X_{k+1} = A X_k + E \tilde{a}_k
\end{equation}

$A$ is the transition matrix, it represents the evolution model.
In our case,
$A = \left( \begin{matrix}
        A_x & 0 \\
        0 & A_y
    \end{matrix}\right)$ and
$A_x = A_y$.
With a timestep $dt$,
$A_x = \left( \begin{matrix}
        1 & dt\\
        0 & 1
    \end{matrix}\right)$.
E is the noise matrix and $\tilde{a}_k = (\tilde{a}_{xk}, \tilde{a}_{yk})^T$ is the noise.
We chose to represent the noise as an acceleration thus
$E = \left( \begin{matrix}
        E_x & 0 \\
        0 & E_y
    \end{matrix}\right)$ and
$E_x = E_y$.
With a timestep $dt$,
$E_x= (\frac{dt^2}{2} , dt)^T$.

The Kalman filter consists in three steps: prediction, innovation, update.
The prediction uses the model to predict the future state estimation $\hat{X}_{k+1|k}$ from the current state estimation
$\hat{X}_{k|k}$.
The covariance matrix $P$ is updated with the model matrix $A$ and with a process noise $Q$.
With the hypothesis that velocity variations are produced by a white noise centered Gaussian acceleration, the process noise $Q$ can be
written $Q =  E Q_a Q_a^T E^T$ with $Q_a$ a square matrix of learned parameters acting as a factorized acceleration
noise matrix.
The observation at time $t_k$ is $Z_k = (x_k, y_k)^T$.
It is matched with the positions from the state vector $X$ with\\
$H = \left( \begin{matrix}
       1 & 0 & 0 & 0 \\
       0 & 0 & 1 & 0
    \end{matrix}\right)$ in the innovation step.

Prediction:
\begin{equation}
    \begin{split}
        \hat{X}_{k+1|k} &= A \hat{X}_{k|k}\\
        P_{k+1|k} &= A P_{k|k} A^T + Q
    \end{split}
    \label{eq_kalman_prediction}
\end{equation}

Innovation:
\begin{equation}
    \begin{split}
        e_{k+1} &= Z_{k+1} - H \hat{X}_{k+1|k}\\
        S_{k+1} &= H P_{k+1|k} H^T + R
    \end{split}
    \label{eq_kalman_innovation}
\end{equation}

Update:
\begin{equation}
    \begin{split}
        K_{k+1} &= P_{k+1|k} H^T S^{-1}_{k+1}\\
    \hat{X}_{k+1|k+1} &= \hat{X}_{k+1|k} + K_{k+1}e_{k+1}\\
    P_{k+1|k+1} &= P_{k+1|k} - K_{k+1} H_{k+1}P_{k+1|k}
    \end{split}
    \label{eq_kalman_update}
\end{equation}

The algorithm~\ref{alg_kalman_prediction} describes the kinematic prediction using the Kalman filter.
For each sample from the training set, a state $\hat{X}_{-15}$ and the covariance matrix $P_{-15}$ are initialized,
then all three steps of the Kalman filter are computed on the 3 seconds history using the parameters
$\sigma_a$ and $R$ and the observations $Z_{k=-14, 0}$.
This allows it to reach the state estimation $\hat{X}_0$ at time $t_0$.
From this time on, only the predictive step is used for prediction with no observation to update it.
The predicted observation for $k \geq 0$ is $\hat{Z}_k = H\hat{X}_{k|0}$.
The associated error covariance matrix estimation is $P^Z_{k} = H P_{k|0} H^T$.
Since there is no new observation after this step, the notation showing the conditioning
to the last observation can be omitted because it is always 0.
As many steps as necessary to fill the desired prediction horizon are made.
We fixed this to 25 steps of 0.2 seconds producing a 5 seconds sequence as prediction.

\begin{algorithm}
    \caption{Kalman prediction}
    \begin{algorithmic}
        \REQUIRE $Q, R$
        \FOR{$Z^i$ in training set}
            \STATE \textbf{Set:} $\hat{X}^i_{-15}, P^i_{-15}$
            \FOR{$k$ from -14 to 0 step 1}
                \STATE $\hat{X}^i_{k}, P^i_{k} \leftarrow \text{Kalman}_{\text{filter}}(Z^i_{k}, \hat{X}^i_{k-1}, P^i_{k-1}, Q, R)$
            \ENDFOR
            \FOR{$k$ from 1 to 25 step 1}
                \STATE $\hat{X}^i_{k}, P^i_{k} \leftarrow \text{Kalman}_{\text{prediction}}(\hat{X}^i_{k-1}, P^i_{k-1}, Q)$
            \ENDFOR
        \ENDFOR
        \RETURN $(\hat{Z}^i_k, P^{Zi}_{k})^{i=1,N}_{k=1, 25}$
    \end{algorithmic}
    \label{alg_kalman_prediction}
\end{algorithm}

The predicted position sequences and the estimated error covariance are compared with the observed next 5 seconds
from the dataset.
The average over the predicted time sequence of the
Negative Log-Likelihood loss (NLL) is to be minimized over a training set from the database.
When averaged over time, it defines a scalar loss function to be minimized.
Otherwise, at each prediction horizon, it used on the test set as a performance indicator.
Performance indicators are defined in the next section.
The Kalman filter function arguments are the current observation and the trainable parameters written
$\operatorname{args} = (\rho, \sigma_a, R, \operatorname{init}) \in ([-1, 1]^4, \mathbb{R}_+^2, \mathbb{S}_+^2, (\mathbb{R}^4, \mathbb{S}_+^4))$,
$\mathbb{S}_+^n$ being the set of symmetric positive definite matrices of size $n\times n$.
The past observations are written $\mathbf{Z}_h = \{Z_k\}_{k=-14,0}$.
The predicted sequence computed using the algorithm~\ref{alg_kalman_prediction} is written
$\text{Kalman}_{\text{pred}}(\mathbf{Z}_h, \operatorname{args})$.
The future observations are written $\mathbf{Z}_f = \{Z_k\}_{k=1, 25}$.

The minimization performed to learn the parameters is expressed as follow:
\[
    \underset{\operatorname{args}}{\operatorname{argmin}} \left( \operatorname{loss}(\text{Kalman}_{\text{pred}}(\mathbf{Z}_h, \operatorname{args}), \mathbf{Z}_f)\right)
\]

The model is implemented with the Pytorch library.
The parameters $\operatorname{args}$ are fitted to the training set
using the Adam optimizer.
%a combination of the RAdam optimizer~\cite{Liu2019}, and the look ahead optimization~\cite{Zhang2019} called ranger and
%available on Github\footnote{\url{https://github.com/lessw2020/Ranger-Deep-Learning-Optimizer}}.
%
Our code for data preprocessing and model training is accessible
on Github\footnote{\url{https://github.com/jmercat/KalmanBaseline}}.

\subsection{Prediction performance indicators}

\subsubsection{Global indicators}

Once the parameters of the model are computed, an evaluation of the predictive performance of the model is made.
In this section, the three most common performance indicators are defined and computed on a test set from the
NGSIM database.
This test set is built using the code from~\cite{Deo2018}.
It is important to note that when used as performance evaluation indicators, the evolution over the predicted horizon
is kept.
The indicators are functions of the prediction step $k$ or as function of time $t$.
The loss to be minimized is a scalar value computed as the average over time of the NLL.

\textbf{The RMSE} computation is made with equation~\eqref{eq_rmse} with $(x_k^i, y_k^i)$ the observed positions and
$(\hat{x}_k^i, \hat{y}_k^i)$ the predicted positions of the $i^{th}$ sequence at time $t_k$.
$N$ is the number of sequences in the subset of the database on which the computation is made.
In our case, it is the number of test sequences which is more than 1.5 million sequences (with sequence overlap).

\begin{equation}
    \text{RMSE}(k) = \sqrt{\frac{1}{N}\sum_{i=1}^N{(x_k^i - \hat{x}_k^i)^2 + (y_k^i - \hat{y}_k^i)^2}}
    \label{eq_rmse}
\end{equation}

\noindent
\textbf{The FDE} computation is made with equations~\eqref{eq_fde}.
\begin{equation}
    \text{FDE}(k) = \frac{1}{N}\sum_{i=1}^N{\sqrt{(x_k^i - \hat{x}_k^i)^2 + (y_k^i - \hat{y}_k^i)^2}}
    \label{eq_fde}
\end{equation}

\noindent
\textbf{The Miss Rate} is the rate with which all proposed predictions miss the final position by more than 2m.
\begin{equation}
    \text{MR}(k) = \frac{1}{N}\sum_{i=1}^N{{\mathds{1}}_{\sqrt{(x_k^i - \hat{x}^{i}_k)^2 + (y_k^i - \hat{y}^{i}_k)^2}>2}}
    \label{eq_mr}
\end{equation}

\noindent
\textbf{The NLL} values reported in articles~\cite{Deo2018, Ju2019, Messaoud2019} are unclear,
they may have chosen a variation of the NLL definition and the dimensionality of the error and covariance
may be meters or feet.
To avoid confusion in our result comparison, the equation~\eqref{eq_mean_nll} using the
NLL formulation from equation~\eqref{eq_gauss_nll} is used and is reported using metric inputs.

By definition, with $f$ a probability density function, the NLL is written:

    \begin{equation}
        \text{NLL} = -\ln\left(f(\mathbf{Z}_f|\mathbf{Z}_h)\right)
        \label{eq_nll}
    \end{equation}

The global NLL over time is not used.
Instead the NLL value for each timestep is computed.
To simplify the notations, the time dependency is not explicitly written with index $k$.
At each timestep $k$, the prediction error is modeled as a bivariate Gaussian probability,\\
thus $f(\mathbf{Z}_f|\mathbf{Z}_h)_k = \mathcal{N}((\mathbf{Z}_f - \hat{\mathbf{Z}}_f)_k, \Sigma_k)$
with
    \[\Sigma_k = P_k =
    \left( \begin{matrix}
        \sigma_x^2 & \rho \sigma_x \sigma_y \\
        \rho \sigma_x \sigma_y & \sigma_y^2
    \end{matrix}\right)
    \]
The errors along $x$ and $y$ axis at time $t_k$ are written $d_x$ and $d_y$.
The coefficients $\sigma_x$, $\sigma_y$, and $\rho$ are identified with the coefficients of $P_k$.
Then the NLL of the prediction at a fixed timestep $k$ is given by equation~\eqref{eq_gauss_nll}.

    \begin{equation}
        \begin{split}
            \text{NLL}(dx, dy, \Sigma) =
            & \frac{1}{2}\underbrace{\frac{1}{(1 - \rho^2)}\biggl(\frac{d_{x}^2}{\sigma_{x}^2} + \frac{d_{y}^2}{\sigma_{y}^2}
            -2 \rho \frac{d_{x} d_{y}}{\sigma_{x} \sigma_{y}}\biggr)
            }_{(Z_k - \hat{Z}_k)^T\Sigma_k^{-1}(Z_k - \hat{Z}_k)}\\
            & + \underbrace{\ln\biggl(\sigma_x\sigma_y\sqrt{1-\rho^2}\biggr)}_{\ln(\sqrt{|\Sigma_k|})}\\
            & + \ln(2\pi)
        \end{split}
        \label{eq_gauss_nll}
    \end{equation}

Over the dataset, the mean NLL (MNLL) value at time $t_k$ is given by equation~\eqref{eq_mean_nll}.

\begin{equation}
    \text{MNLL}(k) = \frac{1}{N}\sum_{i=1}^N{\text{NLL}((x_k^i - \hat{x}_k^i, y_k^i - \hat{y}_k^i, P_k))}
\label{eq_mean_nll}
\end{equation}

\begin{figure*}[ht]
      \centering
%      trim=left botm right top
      \includegraphics[width=\linewidth, clip, trim=2.3cm 6cm 2.5cm 7.4cm]{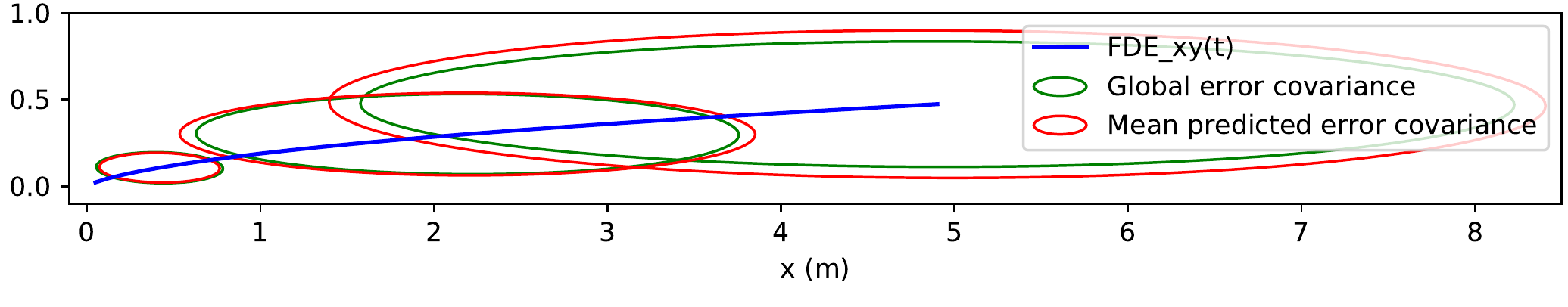}
    \caption{In blue $(\text{FDE}_x(t), \text{FDE}_y(t))$ parametric curve as a function of the prediction time.
            Covariance ellipses at 1s, 3s, and 5s of the predicted sequences.
            Green ellipses represent the prediction error covariance computed over the test examples.
            Red ellipses represent the estimated error covariance of each prediction averaged over the test examples.}
    \label{fig_ellipses}
\end{figure*}

%The optimizations are made with gradient descent over the train dataset to minimize either the NLL loss or the MSE loss.
Results from the computation of these performance indicators over the test set are reported in table~\ref{tab_CV}.
All RMSE and Final Displacement Error (FDE) values are similar to the same indicators from the literature.
However, we obtain a lower NLL value.
This means that all position predictions share the same accuracy but the associated standard deviation estimations are
different.

\subsubsection{Covariance prediction assessment}

A lower NLL value means that on average, when the prediction error is high, the estimated covariance is also
high to lower the first term of equation~\ref{eq_gauss_nll}.
In the same way, when the prediction error is low, the estimated covariance is also low to lower the second term of
equation~\ref{eq_gauss_nll}.
This is a good indicator that the covariance of the prediction error is well estimated on each sample.
However, it is not a very intuitive indicator and it could be lowered either with a better covariance estimation or
a better prediction accuracy.
To assess the prediction error covariance estimation separately from the prediction accuracy and to give an
intuitive representation of it, we also perform a global error covariance analysis.
Figure~\ref{fig_ellipses} shows a combined representation of the FDE as a parametric curve $(\text{FDE}_x(t),
\text{FDE}_y(t))$.
At three times in the prediction sequence, we represent the mean error covariance estimations
, and the matching computed error covariance estimations.
The error covariance matrices are computed at 1s, 3s and 5s of the trajectory predictions over the test set.
They are compared with the mean values of the estimated error covariance matrices.
If the prediction error are unbiased, the average of the estimated error covariance matrices at time $t_k$
is an estimation of the global error covariance matrix at the same time:
\begin{equation}
    \underset{Z_k \in \text{dataset}}{\operatorname{Cov}}(Z_k - \hat{Z}_k) \approx \frac{1}{N}\sum_{i=1}^N{P_k^i}
\end{equation}
%The error caused by averaging covariances with small bias grows linearly with the bias magnitude.
In our case, the error bias is lower than 5\% of the RMSE at all prediction time.
Thus, the average of the error covariances is a good estimation of the predicted overall error covariance.
Using this result, it can be compared with the overall error covariance estimated
from the prediction error over the whole test set.
In figure~\ref{fig_ellipses}, the error covariances are represented as ellipses.
There is a good match between predicted error covariance ellipses and the global
error covariance ellipses.
This does not prove that each prediction error covariance is well estimated but it shows a good overall
covariance estimation.
Combined with low NLL values, this is a satisfying covariance estimation assessement.
%This is also assessed by the low MNLL values in table~\ref{tab_CV}.
%Training with the MSE loss is not sufficient to learn the parameters.
%However, after pre-training with the NLL loss, the model can be retrained with the MSE loss.
%After that training, RMSE values are very slightly lower but as shown in figure~\ref{fig_ellipses_mse}, the
%error covariance is poorly estimated.
%This is also visible in the MNLL values in table~\ref{tab_CV}.

The NGSIM dataset contains perception errors, as shown by the consistency analysis~\cite{Punzo2009}.
Thus, the distance between the predicted trajectories and the observed future is caused
by prediction errors but also by perception errors.
This means that perfect predictions would still produce positive error values.
%To appreciate this, the NLL value considers the estimated variance of the prediction.
The estimated error covariance accounts for both error sources.
Thus the NLL value that accounts for it is a preferable indicator for performance comparisons than RMSE or FDE.
Moreover, the NLL is extendable to other distributions such as Gaussian Mixtures whereas FDE and RMSE are
not good error indicators for multimodal predictions.
For these reasons, having a well-defined baseline correctly estimating its error covariance and its NLL evaluation is
important before producing more complex models.
It is unclear what the reported NLL values from~\cite{Deo2018} mean.
As shown in table~\ref{tab_CV} this would either mean that the estimated likelihood is three order of magnitude worst
than the one we produce with another constant velocity model or, a more likely hypothesis,
that a different computation of the MNLL indicator has been used.

\begin{table}
    \centering
     \caption{Comparison of RMSE, MNLL and FDE results for \textbf{constant velocity models} with the NGSIM test set
    preprocessed with the code published by~\cite{Deo2018} (In~\cite{Xu2019} a different set from NGSIM and 10Hz
    observations instead of 5Hz are used).
     Our results are reported as "ours".
%     Our results obtained from the minimization of the NLL and retraining with
%    the MSE loss are reported as ours$_{\text{NLL}}$ and ours$_{\text{MSE}}$.
     }
    \begin{tabular}{l|cccccc}
        \hline
        \multicolumn{2}{l}{Time horizon}                     & 1s   & 2s   & 3s   & 4s   & 5s   \\
        \hline
        \multirow{3}{*}{RMSE} &From~\cite{Deo2018}           & 0.73 & 1.78 & 3.13 & 4.78 & 6.68 \\
                              &From~\cite{Xu2019}            & 0.48 & 1.50 & 2.91 & 4.72 & NA   \\
                              &ours                          & 0.75 & 1.81 & 3.16 & 4.80 & 6.70 \\
%                              &ours$_{\text{MSE}}$           & 0.75 & 1.81 & 3.16 & 4.79 & 6.69 \\
        \hline
        \multirow{2}{*}{MNLL}  &From~\cite{Deo2018}          & 3.72 & 5.37 & 6.40 & 7.16 & 7.76\\
                               &ours                         & 0.80 & 2.30 & 3.21 & 3.89 & 4.44\\
%                              &ours$_{\text{MSE}}$           & 0.90 & 2.40 & 3.33 & 4.03 & 4.59\\
        \hline
         \multirow{1}{*}{FDE}&ours                          & 0.46 & 1.24 & 2.27 & 3.53 & 4.99\\
        \hline
         \multirow{1}{*}{MR}&ours                           & 0.02 & 0.20 & 0.44 & 0.61 & 0.71\\
        \hline
    \end{tabular}
    \label{tab_CV}
\end{table}

\section{Multi-modal constant velocity prediction}

\begin{figure*}
    \centering
    \includegraphics[width=\textwidth]{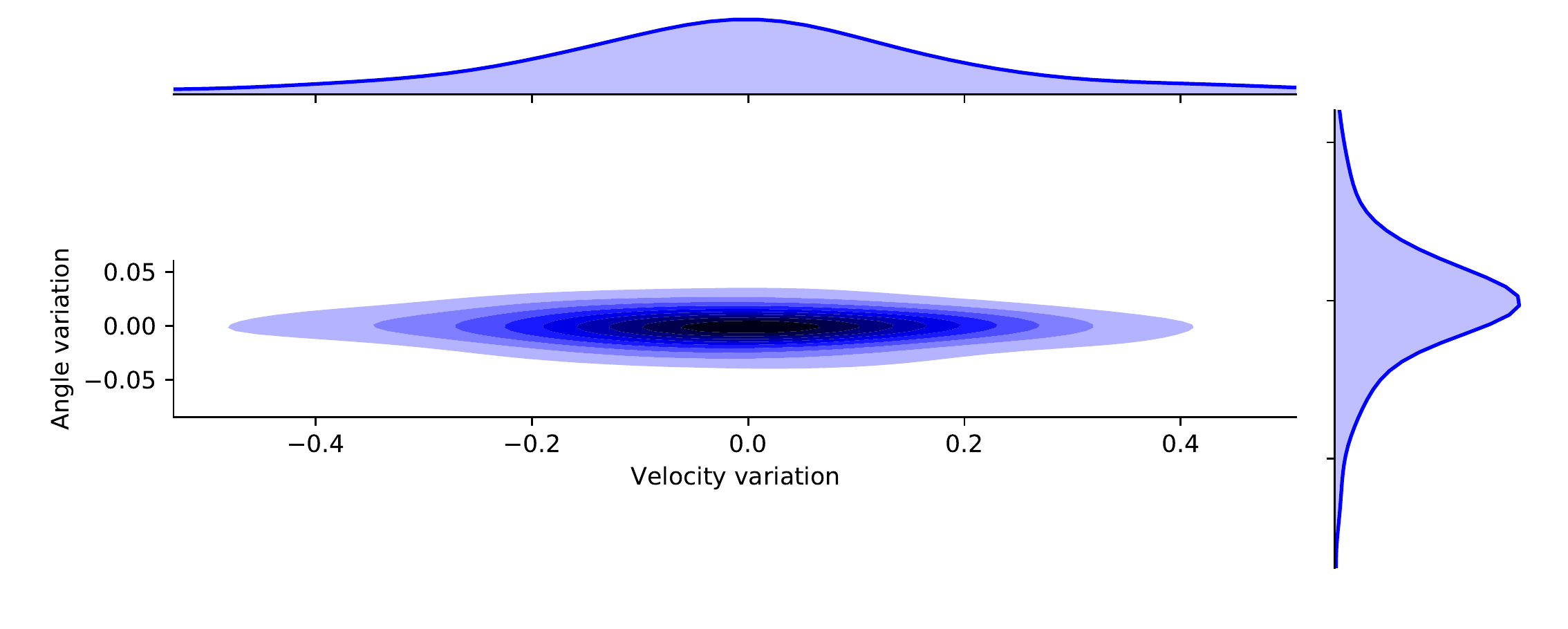}
    \caption{Distribution of velocity and heading angle variation from the estimated values on a random subset of NGSIM.}
    \label{fig:explor_dist}
\end{figure*}

Following the lead of Schöller \textit{et al.}~\cite{Scholler2020}, we want to produce a multi-modal baseline model for vehicle trajectory prediction.
To produce different trajectories, the constant velocity prediction baseline described in the previous section is slightly modified.
We want to explore several possible predictions.
To this end, we define an input standard deviation over the initial heading angle and a standard deviation over a velocity factor.
This defines a bivariate Gaussian that we call the exploration distribution.
We have represented this distribution computed on a subset of the NGSIM dataset in figure~\ref{fig:explor_dist}.
There is no distortion from this distribution to the distribution of error produced on the trajectory, only re-scaling.
We define a new constant velocity prediction for each exploration sample $(\theta, \alpha_v)$ from this distribution.
The modified constant velocity prediction follows these three steps:
Firstly, the initial state of the vehicle is estimated using the Kalman filter.
Then, this state is modified using an exploration sample.
The initial velocity is rotated by $\theta$ and multiplied with a factor $\alpha_v$.
Finally, the constant velocity prediction is produced with the modified initial state.
With this simple procedure, the constant velocity model can be sampled in the same fashion as in~\cite{Scholler2020}.
We vary the velocity and not only the heading angle because with vehicle trajectories, the velocity varies a lot more than with pedestrian trajectories.
Moreover, the heading angle is also less susceptible to change.

We want to explore diverse possibilities using only $k$ exploration samples.
Instead of sampling the exploration distribution, we quantize the exploration distribution with a k-means algorithm.
The $k$ resulting centroids $c_1, ..., c_k$ are used as exploration samples.
%With $k$ samples, the centroid probability estimation $p_j$ is estimated as the normalized value of the exploration probability density function evaluated at each centroid:
%\begin{equation*}
%    \begin{split}
%        G_\text{PDF}(x, \Sigma) &= \frac {1} {(2\pi)^{N/2} \left|\Sigma\right|^{1/2}}\;\; e^{-\frac{1}{2}( x - \mu)^\top \Sigma^{-1} (x - \mu)}\\
%         p_j &= \frac{G_\text{PDF}(c_j, \Sigma)}{\sum_{i=1}^k G_\text{PDF}(c_i, \Sigma)}
%    \end{split}
%\end{equation*}
With $k$ samples, the $j^\text{th}$ centroid probability estimation $p_j$ is the probability of the Voronoi cell associated to $c_j$ for the exploration distribution represented with the centered Gaussian PDF $G_\text{PDF}$ with covariance $\Sigma$:
\begin{equation*}
        V_\text{cell}(c_j) = \{ z \in \mathbb{R}^2 \  /\  \forall i \neq j \in \llbracket 1, k \rrbracket ~  \| z - c_j \| \le \| z - c_i \| \}
\end{equation*}
\begin{equation*}
        p_j = p(V_\text{cell}(c_j)) = \int_{V_\text{cell}(c_j)}  G_\text{PDF}(z, \Sigma) dr
\end{equation*}
Since the Voronoi cells form a partition of $\mathbb{R}^2$, $\sum_{i=1}^kp_s = 1$.

The constant velocity prediction also predicts an error covariance matrix.
When several modes are considered, each mode represents a part of the distribution described by this error covariance.
Thus, the error covariance around each mode is expected to be lower than the error covariance for one central mode.
We compute a covariance coefficient for each mode as the fraction of the standard deviation inside the Voronoi cell and the exploration standard deviation:
\begin{equation*}
    \alpha_j = \sqrt{\frac{\mathbb{E}_{V_\text{cell}(c_j)}\left[ \left( X - \mathbb{E}_{V_\text{cell}(c_j)}[X] \right)^2 \right]}{\mathbb{E}\left[ X^2 \right]}}
\end{equation*}
Then, the constant velocity prediction with optimized parameters can be used directly to produce multi-modal predictions.
In the following we may refer to the multi-modal prediction as multiple predictions by considering each mode as a prediction proposition.

\subsection{Multi-modal evaluation metrics}

\textbf{The RMSE and FDE} definitions can be adapted for the evaluation of multi-modal predictions by considering that each prediction mode is a different prediction propositions for the same past observations.
When several trajectories with the associated probabilities are produced,
three estimations of the RMSE and FDE are given for a set of $n_\text{mix}$ proposed trajectories:

\noindent
$\bullet$ The prediction with the maximum estimated probability:
{\footnotesize
\begin{equation*}
    \begin{split}
        \text{RMSE}(k) &= \sqrt{\frac{1}{N}\sum_{i=1}^N{\sum_{m=1}^{n_\text{mix}} \mathds{1}_{p_m^{(i)}=p_\text{max}^{(i)}} \left( (\tilde{x}_k^{(i)} - \hat{x}_{k,m}^{(i)})^2 + (\tilde{y}_k^{(i)} - \hat{y}_{k,m}^{(i)})^2\right)}}\\
        \text{FDE}(k) &= \frac{1}{N}\sum_{i=1}^N{\sum_{m=1}^{n_\text{mix}} \mathds{1}_{p_m^{(i)}=p_\text{max}^{(i)}} \sqrt{(\tilde{x}_k^{(i)} - \hat{x}_{k,m}^{(i)})^2 + (\tilde{y}_k^{(i)} - \hat{y}_{k,m}^{(i)})^2}}
    \end{split}
\end{equation*}}
\noindent
$\bullet$ A weighted average of the error for each proposition:
{\footnotesize
\begin{equation*}
    \begin{split}
        p\text{RMSE}(k) &= \sqrt{\frac{1}{N}\sum_{i=1}^N{\sum_{m=1}^{n_\text{mix}} p_m^{(i)} \left( (\tilde{x}_k^{(i)} - \hat{x}_{k,m}^{(i)})^2 + (\tilde{y}_k^{(i)} - \hat{y}_{k,m}^{(i)})^2\right)}}\\
        p\text{FDE}(k) &= \frac{1}{N}\sum_{i=1}^N{\sum_{m=1}^{n_\text{mix}} p_m^{(i)} \sqrt{(\tilde{x}_k^{(i)} - \hat{x}_{k,m}^{(i)})^2 + (\tilde{y}_k^{(i)} - \hat{y}_{k,m}^{(i)})^2}}
    \end{split}
\end{equation*}}
$\bullet$ The error for the trajectories that produces the minimum final displacement error:
{\footnotesize
\begin{equation*}
    \begin{split}
        \text{minRMSE}(k) &= \sqrt{\frac{1}{N}\sum_{i=1}^N{(\tilde{x}_k^{(i)} - \hat{x}_{k, \text{min}}^{(i)})^2 + (\tilde{y}_k^{(i)} - \hat{y}_{k, \text{min}}^{(i)})^2}}\\
        \text{minFDE}(k) &= \frac{1}{N}\sum_{i=1}^N{\sqrt{(\tilde{x}_k^{(i)} - \hat{x}_{k, \text{min}}^{(i)})^2 + (\tilde{y}_k^{(i)} - \hat{y}_{k, \text{min}}^{(i)})^2}}
    \end{split}
\end{equation*}}
This last evaluation requires the knowledge of the future trajectory to select the mode whose last position is the closest to the last future position.
This leads to an overly optimistic error estimation but it is informative about the capacity of the model to fit the different modes.
The weighted average and the maximum probability are overly pessimistic because they may include the error of far-off trajectories that might have been.
The observed future is not always the most probable outcome that could have happened.
Moreover, it mixes the capacity to estimate the probabilities of the different modes and the capacity to fit them.
Thus, these indicators bound the error but are not sufficient to characterize it.

Several prediction propositions can be quasi-superposed.
If each component describes a different mode, the output represents a broader exploration of the less likely modes, and it becomes easier to interpret.
Thus, we propose a similarity indicator for the prediction propositions.
A low similarity indicates a proper differentiation of the prediction propositions.
Thus, for similar values of the performance indicators, a lower similarity is preferable.

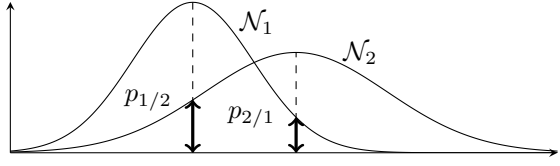
\begin{figure}[H]
    \centering
    \begin{tikzpicture}
     \begin{axis}[
            x=0.8cm,
            y=5cm,
            xmin=1,
            xmax=10,
            ymin=0,
            ymax=0.4,
            axis lines=center,
            ticks=none,
%    %        xlabel={$k_t$},
%    %        ylabel={$f(k_t)$},
%    %        xlabel style={below right},
%    %        ylabel style={above left},
%            % (moved common `addplot' options here)
            smooth,
            domain=0:10,
           samples=101,
            no markers,
        ]

%    % Draw curves
    \addplot [thin, smooth, name path global=first] {dnorm(4,1)};
    \addplot [thin, smooth, name path global=second] {dnorm(5.7,1.5)};

    % Draw vertical line:
    \path [name path global=first_line] ({rel axis cs:0,0}-|{axis cs:4,0}) -- ({rel axis cs:0,1}-|{axis cs:4,0});
    \path [name path global=second_line] ({rel axis cs:0,0}-|{axis cs:5.7,0}) -- ({rel axis cs:0,1}-|{axis cs:5.7,0});

    \path [name intersections={of=first_line and first}] (intersection-1) circle
                coordinate (m1);
    \path [name intersections={of=second_line and second}] (intersection-1) circle
                coordinate (m2);

    \node at ([shift={(6, 4)}]m1) [below right = 0mm and 5mm of m1] {$\mathcal{N}_1$};
    \node at ([shift={(6, 5)}]m2) [right = 5mm of m2] {$\mathcal{N}_2$};
%    \draw [thick](m1) node [dot] {}
%                -- (m1) node[dot, label:] {};
%    \draw (m2) node[dot, label:right:$\mathcal{N}_2$] {};

    \path [name intersections={of=first_line and second}] (intersection-1) circle
                coordinate (p1);
    \path [name intersections={of=second_line and first}] (intersection-1) circle
                coordinate (p2);

    \draw [dashed] (0, 0 -|  m1) node{}
        -- (m1)                node {};
   \draw [dashed] (0, 0 -|  m2) node {}
        -- (m2)                node {};
    \draw [very thick, <->] (0, 0 -|  p1) node{}
        -- (p1)                node [label=left:$p_{1/2}$] {};
    \draw [very thick, <->] (0, 0 -|  p2) node {}
        -- (p2)                node [label=left:$p_{2/1}$] {};
    %\draw [thick] (p2 -| 0, 0) node [dot, label=p1]{};
    %\draw [thick, name intersections={of={second_line and first}}] ({rel axis cs:0,0}-|intersection-1) -- ({rel axis cs:0,1}-|intersection-1);
    \end{axis}
    \end{tikzpicture}

    \caption{Graph of two Gaussian probability density functions (PDF). $p_{1/2}$ and $p_{2/1}$ are the values of one PDF computed
    at the mean value of the other.}
    \label{plt_gauss}
\end{figure}

\textbf{The similarity indicator} is computed as the average probability density of one component center for the other components at the final time.
The figure~\ref{plt_gauss} is a representation of the intermediate computations used to compute the similarity indicator.
For each pair of Gaussian mixture components $i$, and $j$, with $i \neq j$, the probability density function of one component is computed at the centroid (the position of the mean value) of the other.
\pgfmathdeclarefunction{dnorm}{2}{%
  \pgfmathparse{1/(#2*sqrt(2*pi))*exp(-((x-#1)^2)/(2*#2^2))}%
}
This produces the two values $p_{i/j}$ and $p_{j/i}$.
The similarity indicator is defined in equation~\eqref{eq_sim_loss}.

\begin{equation}
    \text{SIM}(k) = \frac{1}{n_{\text{mix}}(n_{\text{mix}}-1)}\sum_{i=1}^{n_\text{mix}}\sum_{j \neq i}^{n_\text{mix}} p_{i/j}(k)p_{j/i}(k)
    \label{eq_sim_loss}
\end{equation}

\textbf{The NLL} is easily extended to a Gaussian mixture by computing the "log-sum-exp" of the individual prediction NLLs:
\begin{equation}
    \text{NLL}_{mm} = -\log\left(\sum_{i=1}^{\text{mix}} p_i \exp(-NLL_i)\right) 
\end{equation}

\textbf{The Miss Rate} is extended to multiple predictions by considering the rate at which none of the proposed predictions comes closer than 2m to the observed future.

\subsection{Application}
For a prediction $f(\mathbf{x}) = (\mathbf{y}, \Sigma)$, the associated multi-modal prediction is $f(x_i) = (y_i, \alpha_i\Sigma, p_i)$.
It uses the same prediction function $f$ that leads to the same error covariance $\Sigma$ independently of $\mathbf{x}$.
Therefore, this fully defines a multi-modal constant velocity prediction in the form of a Gaussian mixture.
In the table~\ref{tab:results_cv_mm}, we present the results for a constant velocity prediction with $k=6$ modes exploring a distribution defined with $\sigma_\theta = 0^\circ$ and $\sigma_{\alpha_v} = 10\%$.
No training is necessary to compute the results with a new set of parameters.
Thus we obtained these parameters with a simple grid search over $\sigma_\theta$, and $\sigma_{\alpha_v}$ using a fixed number of samples $k=6$.
The results show a sharp decrease in the miss rate showing that the modes explore a state-space that fits the data.
The best exploration angle is $0^\circ$.
It shows how little effect the lateral maneuvers have on the NGSIM dataset while the velocity varies more.
However, the similarity values are high.
It means that this model is elongating the velocity distribution more than it is producing new modes.
The NLL at 5s is 4.44 for one mode and 5.39 with the 6 modes.
This could probably be improved with a better estimation of the covariance coefficients $\alpha$.  % or the probability associated to each mode.
These parameters could also be optimized using the training data.
However, besides the NLL values, this is a satisfying baseline for multi-modal prediction.
Remarkably, the miss rate is at 30\%, while the same number of modes with the Convolutional Social Pooling from~\cite{Deo2018} shows a 44\% miss rate.
This result is a strong argument against using hand-defined maneuvers as modes.

\begin{table}[h]
\centering
\caption{Evaluation of the multi-modal constant velocity prediction with $k=6$, $\sigma_\theta = 0^\circ$ and $\sigma_{\alpha_v} = 10\%$}
\begin{tabular}[h]{l|cccccc}
    \hline
    Time horizon  & 1s   & 2s   & 3s   & 4s   & 5s   \\
    \hline
    \multirow{1}{*}{pRMSE (m)} & 1.26 & 2.72 & 4.40 & 6.30 & 8.40 \\
    \hline
    \multirow{1}{*}{RMSE (m)} & 0.88 & 2.04 & 3.49 & 5.21 & 7.15 \\
    \hline
    \multirow{1}{*}{minRMSE (m)} & 0.53 & 1.18 & 1.96 & 2.98 & 4.18 \\
    \hline
     \multirow{1}{*}{pFDE (m)} & 1.02 & 2.23 & 3.65 & 5.23 & 6.99\\
    \hline
     \multirow{1}{*}{FDE (m)} & 0.52 & 1.32 & 2.39 & 3.67 & 5.14\\
     \hline
     \multirow{1}{*}{minFDE (m)} & 0.26 & 0.58 & 1.02 & 1.59 & 2.28\\
    \hline
    \multirow{1}{*}{NLL (m)} & 2.45 & 3.35 & 4.12 & 4.81 & 5.39\\
    \hline
     \multirow{1}{*}{MR} & 0.01 & 0.03 & 0.10 & 0.20 & 0.30\\
    \hline
    \multirow{1}{*}{Sim} & 12.92 & 1.26 & 0.36 & 0.18 & 0.12\\
\end{tabular}
\label{tab:results_cv_mm}
\end{table}

\section{Model based RNN predictions}

In this section, we extend the constant velocity model with a prediction model that depends on the acceleration and
a recurrent neural network for command prediction.
The predicted command is a variation of the acceleration which can be interpreted as a jerk.
This relies on a Kalman filter using a state with velocities and accelerations
$X=(x, v_x, a_x, y, v_y, a_y)^T$.

The evolution of the state is written as follow:
\begin{equation}
    X_{k+1} = A X_k + B u_k + E \tilde{j_k}
    \label{eq_kalman_command}
\end{equation}

with the transition matrix
\begin{eqnarray*}
    A = \left( \begin{matrix}
        A_x & 0 \\
        0 & A_y
    \end{matrix}\right)\\
    A_x = A_y = \left( \begin{matrix}
        1 & dt & dt^2/2 \\
        0 & 1  & dt \\
        0 & 0  &  1
    \end{matrix}\right)
\end{eqnarray*}
We write $\tilde{j}_k = (\tilde{j}_{xk}, \tilde{j}_{yk})^T$ as the noise jerk vector.
It is modeled as a zero-mean Gaussian random variable.
Its estimated standard-deviation is written $\sigma_{\hat{u}k}$.
The command matrix $B$ and the noise matrix $E$ are equal because the
command and the noise are both jerk.

\[B = E = \left( \begin{matrix}
        dt^3/6 & dt^2/2 & dt & 0 & 0 & 0\\
        0 & 0 & 0 & dt^3/6 & dt^2/2 & dt
\end{matrix}\right)^T
\]

Thus the evolution of the state simplifies to:

\begin{equation}
    X_{k+1} = A X_k + B (u_k + \tilde{j_k})
    \label{eq_kalman_command2}
\end{equation}

The Kalman filter with recurrent command prediction consists of the three Kalman steps
with an extra line for the command prediction in the prediction step written in equations~\eqref{eq_kalman_lstm_prediction}.
This new line defines a function with two outputs: $h_{k} = (\hat{u}_{k}, q_{\hat{u}k})$, and $c_k$.
$h_k$ contains the command and its variance, $c_k$ is a recurrent value that serves as memory
from one timestep to the next.
All three vectors are defined with a single equality in equation~\eqref{eq_kalman_lstm_prediction}.

New Prediction:
\begin{equation}
    \begin{split}
        (\hat{u}_{k}, q_{\hat{u}k}), c_k &= \operatorname{RNNCell}(\hat{X}_{k|k}, (\hat{u}_{k-1}, q_{\hat{u}k-1}), c_{k-1}) \\
        \hat{X}_{k+1|k} &= A \hat{X}_{k|k} + B \hat{u}_{k}\\
        P_{k+1|k} &= A P_{k|k} A^T + B q_{\hat{u}k}q_{\hat{u}k}^T B^T
    \end{split}
    \label{eq_kalman_lstm_prediction}
\end{equation}

Innovation and update steps are written the same way as equations~\eqref{eq_kalman_innovation}
and~\eqref{eq_kalman_update}.
For the RNNCell function we chose an LSTM cell.
Its computation is written in equation~\eqref{eq_lstm}, with the output $h_{k} = (\hat{u}_{k}, q_{\hat{u}k})$.
The symbol "$\odot$" denotes elementwise multiplication.
The functions $\sigma: x \rightarrow \frac{1}{1+e^{-x}}$ and $\tanh$ are applied elementwise.
The LSTM cell depends on weights
$W^i, W^f, W^o, W^g, U^i, U^f, U^o, U^g$ and bias $b_i, b_f, b_g, b_o$.
These parameters are fitted in the loss minimization process.
\begin{equation}
%\begin{array}{r@{}l}
    \begin{split}
        f_{k} &=\sigma\big(x_{k}U^{f}+h_{k-1}W^{f} +b_f\big)\\
        i_{k} &=\sigma\big(x_{k}U^{i} +h_{k-1}W^{i}+b_i\big)\\
        g_{k} &=\tanh\big(x_{k}U^{g}+h_{k-1}W^{g}+b_g\big)\\
        o_{k} &=\sigma\big(x_{k}U^{o}+h_{k-1}W^{o}+b_o\big)\\
%    s_k\begin{cases}
%        \begin{aligned}
        c_{k} &= f_k \odot c_{k-1} + i_k \odot g_k\\
        h_{k} &= o_{k} \odot \tanh(c_{k})  
%        \end{aligned}
%    \end{cases}\nonumber
    \end{split}
%\end{array}
\label{eq_lstm}
\end{equation}

Exactly as before, predictions are produced with the algorithm~\ref{alg_kalman_prediction} however
they are computed with this new Kalman filter.
Another modification is that the predicted commands are computed with the RNNCell but not used during the
filtering phase.
This allows the initialization of the state estimation and recurrent parameters of the RNNCell without
demanding to the command predictor to be robust to the initially bad estimations of the state.
Then, during the prediction phase, the commands are computed and used.
The model parameters are learned on the training set with the minimization of the NLL loss.

\begin{table}
    \centering
    \caption{Comparison of RMSE and NLL results for single learned dynamic models observing the trajectory of a single
    vehicle without context nor interaction.
    The NGSIM test set preprocessed with the code published by~\cite{Deo2018} is used.
    *Results from our training of their model.}
    \begin{tabular}{l|cccccc}
        \hline
        \multicolumn{2}{l}{Time horizon}               & 1s   & 2s   & 3s   & 4s   & 5s   \\
        \hline
        \multirow{2}{*}{RMSE} &V-LSTM~\cite{Deo2018}*  & 0.67 & 1.62 & 2.86 & 4.40 & 6.26 \\
                              &ours                    & 0.68 & 1.54 & 2.72 & 4.20 & 5.95 \\
        \hline
        \multirow{2}{*}{MNLL}  &V-LSTM~\cite{Deo2018}* & -0.02 & 1.64 & 2.59 & 3.26 & 3.79\\
                              &ours                    & -0.15 & 1.50 & 2.50 & 3.19 & 3.73\\
        \hline
        \multirow{2}{*}{FDE}  &V-LSTM~\cite{Deo2018}* & 0.42 & 1.14 & 2.10 & 3.28 & 4.66\\
                              &ours                    & 0.40 & 1.09 & 2.02 & 3.19 & 4.56\\
        \hline
        \multirow{2}{*}{MR}  &V-LSTM~\cite{Deo2018}* & 0.01 & 0.17 & 0.41 & 0.59 & 0.71\\
                              &ours                    & 0.01 & 0.15 & 0.39 & 0.58 & 0.71\\
        \hline
    \end{tabular}

    \label{tab_dyn}
\end{table}

We obtain the results shown in table~\ref{tab_dyn} compared with results from the V-LSTM prediction from~\cite{Deo2018}
that is made with an LSTM encoder and LSTM decoder (we recomputed the results to harmonize the NLL definition).
This shows that forcing this structure is not detrimental and even somewhat better for RMSE performances and
estimation of the covariance.
Using this, interpretations, and constraints such as maximum acceleration or velocity recommendations are
easy to produce.
Moreover, the method should usable with any Kalman model in the generic form of
equations~\eqref{eq_kalman_prediction},~\eqref{eq_kalman_innovation},~\eqref{eq_kalman_update}.
However, in our attempts to use it with a bicycle model, the optimization process became unstable, often leading
to out of bound values or invalid results.
With a careful initialization and a scheduled training process, results were obtained but the error estimation did
not fit as well as with the simpler model presented here.

In~\cite{Ju2019}, a Kalman filter using predicted accelerations from a neural network is also used.
However, acceleration predictions and Kalman filtering are separated.
The neural network takes observations as input and produces the acceleration prediction sequence.
This sequence is then fed to the Kalman filter.
In contrast, our method takes advantage of the whole kinematic state evaluation to predict the commands at each step.
In~\cite{Coskun2017}, another combination of LSTM and Kalman filter is made.
They replaced the state update with the LSTM cell (intuitively, instead of $X_{k+1} = A X_k$ they use
$X_{k+1} = \text{LSTM}(X_k)$).
In our case, the kinematic model forces the trajectory to have inertia which is known to play an important part.
It keeps a kinematic interpretation of the state $X$ and does not require the LSTM Jacobian matrix computation.

\section{CONCLUSIONS}

We have established reproducible baselines results for simple prediction methods on the NGSIM dataset.
NLL results were lower than comparable published baselines
and should push further work toward better estimation and validation of prediction error covariance.
A global covariance estimation assessment has been presented with superposition of ellipses from
computed prediction error covariance knowing the future estimations
and mean estimated error covariance at prediction time.
We proposed an extension of the constant velocity model for multi-modal predictions that can serve as a baseline for more advanced work that still compare their results with the uni-modal constant velocity model.
Finally, an easy to implement extension of the constant velocity model allowing command prediction was produced.
It gives a baseline for machine learning trajectory prediction models with a good error covariance estimation.

\subsubsection*{Acknowledgements} We are grateful to Edouard Leurent for his comments and corrections.

%Error covariance estimations using the presented models are an improvement over published baseline models.

%%%%%%%%%%%%%%%%%%%%%%%%%%%%%%%%%%%%%%%%%%%%%%%%%%%%%%%%%%%%%%%%%%%%%%%%%%%%%%%%

\end{document}